\documentclass[letterpaper, 10 pt, conference]{IEEEtran}

\newif\ifarxiv
\arxivtrue
\def\MYTITLE{Evaluation of Mobile Environment for Vehicular Visible Light Communication Using Multiple LEDs and Event Cameras}

\ifarxiv
\usepackage[breaklinks,colorlinks]{hyperref}
\hypersetup{
  pdftitle={\MYTITLE},
  pdfsubject={Computer Vision, Robotics},
  pdfauthor={Ryota Soga, Takaya Yamazato, Shan Lu, Shintaro Shiba, Quan Kong, Norimasa Kobori, Tsukasa Shimizu},
  pdfkeywords={Event Cameras, Visible Light Communication, Optical Communication}
}
\usepackage[absolute]{textpos}
\else
\usepackage[pagebackref,breaklinks,colorlinks]{hyperref}
\fi

\IEEEoverridecommandlockouts
\usepackage[T1]{fontenc} 
\usepackage{cite}
\usepackage{amsmath,amssymb,amsfonts}
\usepackage{algorithmic}
\usepackage{graphicx}
\usepackage{textcomp}
\usepackage{xcolor}
\usepackage{subcaption}
\usepackage{ragged2e}
\usepackage{url}
\usepackage{setspace}
\usepackage{amsmath}
\usepackage{orcidlink}
\usepackage[capitalize]{cleveref}
\crefname{section}{Sec.}{Secs.}
\Crefname{section}{Section}{Sections}
\Crefname{figure}{Figure}{Figures}
\crefname{figure}{Fig.}{Figs.}
\Crefname{table}{Table}{Tables}
\crefname{table}{Tab.}{Tabs.}

\begin{document}

\ifarxiv
\definecolor{somegray}{gray}{0.5}
\newcommand{\darkgrayed}[1]{\textcolor{somegray}{#1}}
\begin{textblock}{11.5}(2.25, 0.3)  % {hsize}(hpos,vpos)
\begin{center}
\darkgrayed{This paper has been accepted for publication at the IEEE Intelligent Vehicles Symposium (IV), \\
Napoca, Romania, 2025.
\copyright IEEE}
\end{center}
\end{textblock}
\fi

\title{\MYTITLE}

%要メールアドレス、所在地の確認
\author{
\hspace*{0cm}  % 左方向へのスペース調整
\IEEEauthorblockN{
Ryota Soga$^{1}$, Shintaro Shiba$^{2}$, Quan Kong$^{2}$, Norimasa Kobori$^{2}$,\\}
\IEEEauthorblockN{Tsukasa Shimizu$^{3}$, Shan Lu$^{1}$, Takaya Yamazato$^{1}$\\}
\IEEEauthorblockA{$^{1}$School of Engineering, Nagoya University,
Nagoya, Japan\\
\{soga, lu, yamazato\}@yamazato.nuee.nagoya-u.ac.jp\\}
\IEEEauthorblockA{$^{2}$Woven by Toyota, Inc.,
Tokyo, Japan\\
\{shintaro.shiba, quan.kong, norimasa.kobori\}@woven.toyota\\}
\IEEEauthorblockA{$^{3}$TOYOTA MOTOR CORPORATION,
Toyota, Japan\\
tsukasa\_shimizu@mail.toyota.co.jp}
}

\maketitle

\begin{abstract}
In the fields of Advanced Driver Assistance Systems (ADAS) and Autonomous Driving (AD), sensors that serve as the ``eyes'' for sensing the vehicle's surrounding environment are essential. Traditionally, image sensors and LiDAR have played this role. However, a new type of vision sensor, event cameras, has recently attracted attention. Event cameras respond to changes in the surrounding environment (e.g., motion), exhibit strong robustness against motion blur, and perform well in high dynamic range environments, which are desirable in robotics applications.
Furthermore, the asynchronous and low-latency principles of data acquisition make event cameras suitable for optical communication.
% This study explores the addition of communication functionality to event cameras.
By adding communication functionality to event cameras, it becomes possible to utilize I2V communication to immediately share information about forward collisions, sudden braking, and road conditions, thereby contributing to hazard avoidance. Additionally, receiving information such as signal timing and traffic volume enables speed adjustment and optimal route selection, facilitating more efficient driving.
In this study, we construct a vehicle visible light communication system where event cameras are receivers, and multiple LEDs are transmitters. In driving scenes, the system tracks the transmitter positions and separates densely packed LED light sources using pilot sequences based on Walsh-Hadamard codes. As a result, outdoor vehicle experiments demonstrate error-free communication under conditions where the transmitter-receiver distance was within 40 meters and the vehicle's driving speed was 30 km/h (8.3 m/s).
\end{abstract}

\begin{figure}[ht]
    \centering
    \includegraphics[width = \linewidth]{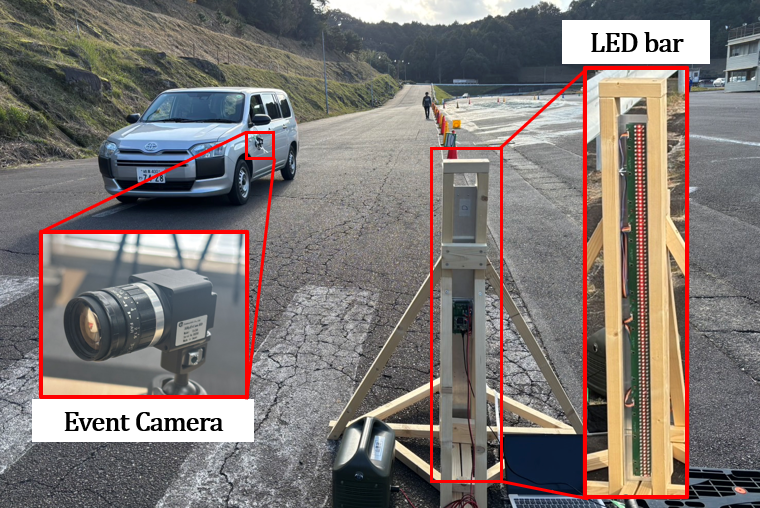}
    \caption{A Vehicular Visible Light Communication System Using an Event Camera}
    \label{fig:rec-and-tra}
\end{figure}

\section{Introduction}
\label{sec:introduction}
In autonomous driving systems, sharing information with other vehicles and road infrastructure is essential~\cite{IEEE2407, IEEE9307}. The simplest method of such information sharing is wireless communication using radio frequency (RF). However, many devices, such as mobile communication systems and TVs, utilize radio waves. To enable a large number of vehicles to interconnect via RF for Intelligent Transportation Systems (ITS) purposes, it is necessary to allocate frequency bands~\cite{IEEE2019}. As a result, the frequency bands may become congested, leading to a decline in speed and instability in data transmission and reception, potentially hindering autonomous driving.

\begin{figure}[t]
    \centering
    \includegraphics[width = \linewidth]{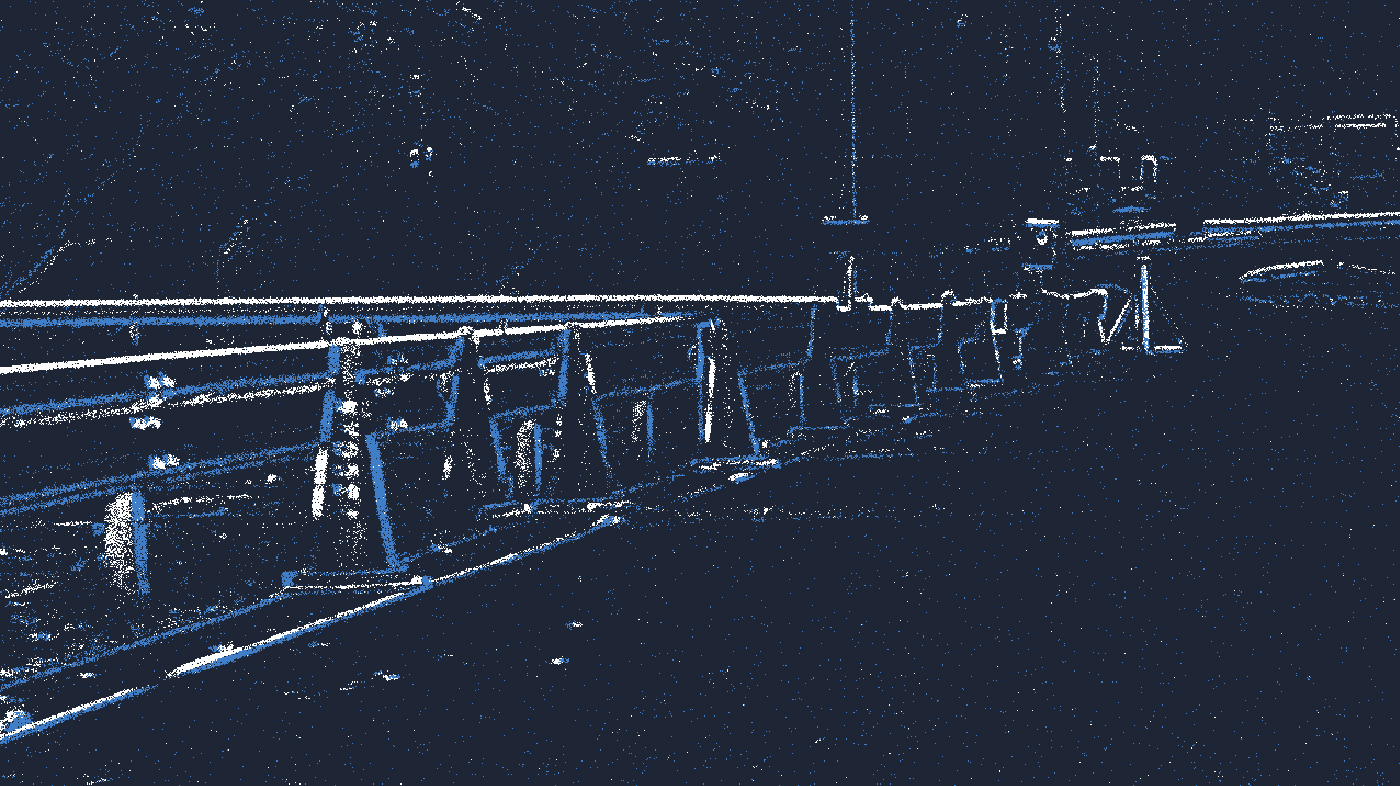}
    \caption{Shows an image frame created by compiling 20 ms of event data captured by an in-vehicle event camera}
    \vspace{-4mm}
    \label{fig:eventdata}
\end{figure}

To solve this challenge, vehicular visible light communication (VLC) has gained attention. VLC does not consume radio frequency bandwidth and enables V2X communication. Furthermore, methods combining RF and VLC are being explored to achieve both high reliability and low latency~\cite{IEEE2300}.
So far, photodiodes (PDs) and image sensors have been commonly used as receivers for VLC. Each has its advantages and disadvantages: PDs achieve high communication rates due to their high temporal resolution~\cite{IEEE2022}. However, they face challenges in separating multiple light sources and resisting external interference, such as sunlight. In contrast, image sensors offer high spatial resolution, enabling the separation of multiple light sources. Since they are conventional sensors for perception in AD/ADAS systems, existing cameras such as in-vehicle ones can be used directly~\cite{IEEE2024}.
Furthermore, when combined with visible light positioning (VLP), it can be used for ranging~\cite{Katayama2023}. However, the temporal resolution of conventional image sensors is typically limited to approximately 16 ms (i.e., 60 fps), and it is challenging to realize high communication rates alone. High-speed image sensors suffer from large data that severely constraints the system bandwidth, yet their temporal resolution is limited to around 1 ms.
In this study, we propose a vehicular visible light communication system using an ``event camera'' as the receiver to address these challenges (Fig.~\ref{fig:rec-and-tra}) Event cameras asynchronously record only pixels where intensity changes occur, outputting information on polarity, pixel coordinates, and timestamp, thereby achieving high temporal resolution while reducing data size~\cite{Lichtsteiner2006isscc}. Like conventional image sensors, event cameras also serve as ``eyes'' for recognizing the surrounding environment for autonomous robots and vehicles~\cite{Gallego2022pami,Gehrig2024nature}.
Fig.~\ref{fig:eventdata} shows an example image created by integrating 20 ms of data captured by an in-vehicle event camera. Previous studies have demonstrated high communication rates using event cameras as receivers, but most evaluations have been conducted in controlled environments where the camera is static~\cite{IEEE2018, ACM2019, Arxiv2024}. Vehicular applications require event cameras to receive and analyze signals in dynamic environments.
% adaptation to dynamic environments is required.

In a mobile environment, the key challenges is tracking the transmitter. Additionally, if it becomes possible to individually track and separate multiple information sources, different data can be transmitted, thereby improving the communication rate. To address them, this study proposes an LED position identification method using pilot sequences (known signals) based on Walsh-Hadamard (WH) codes. We assign unique WH codes to each LED and insert pilot sequences between transmitted data. The receiver side (i.e., event cameras) calculates correlations with each WH code for each grid and estimates the probability of LED presence within each grid. This allows for LED position identification regardless of orientation or configuration, and the interference from other LEDs is mitigated by the orthogonality of WH codes.
Additionally, we find that vehicle vibrations caused by road surfaces can result in movements that cannot be tracked during pilot sequence transmission intervals. To address this, we propose movement estimation and correction between pilot sequences. 

The evaluations in outdoor driving scenes demonstrate error-free communication at speeds of up to 30 km/h (8.3 m/s) and distances of up to 40 m, assuming small mobility use cases. We find that the achieved communication rate during these conditions is 28 kbps.
The contributions of this study can be summarized as follows:
\begin{itemize}
    \item By employing a pilot sequence, transmitter tracking become possible, enabling the first adaptation of event camera-based VLC to in-vehicle environments.
    \item The same pilot sequence-based approach also allowed the separation of densely packed multiple LEDs, enabling MISO communication and significantly improving the communication rate.
    \item This study presents the first experimental demonstration of event camera-based VLC under actual vehicle-mounted conditions. The results show that error-free communication was successfully achieved at a speed of 30 km/h (8.3 m/s) within a communication distance of less than 40 meters.
\end{itemize}
\section{System Model}
\label{sec:system_model}
\begin{figure*}[tbp]
\centering
\includegraphics[width=0.95\linewidth]{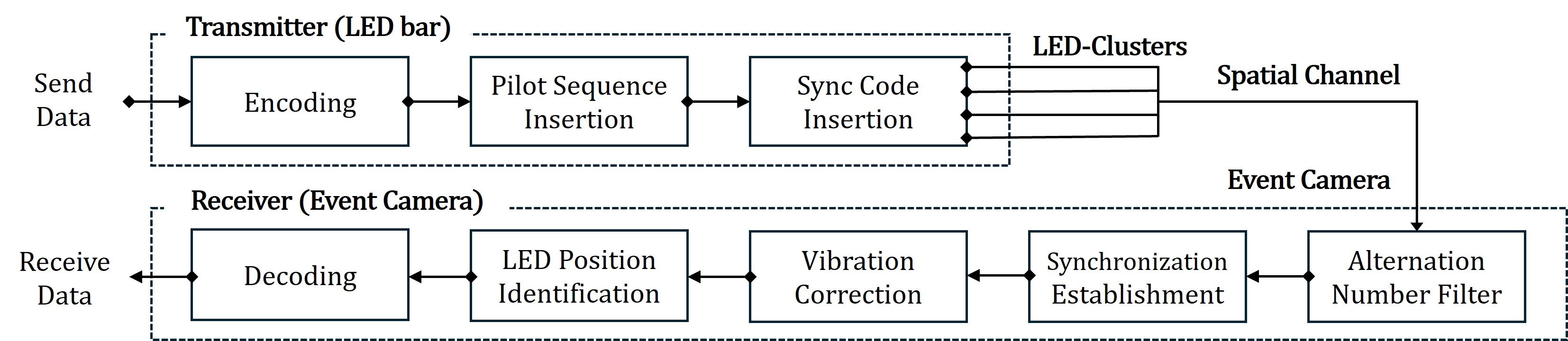}
\caption{System model}
\vspace{-4mm}
\label{fig:systemmodel}
\end{figure*}
\begin{figure}[ht]
    \centering
    \includegraphics[width = \linewidth]{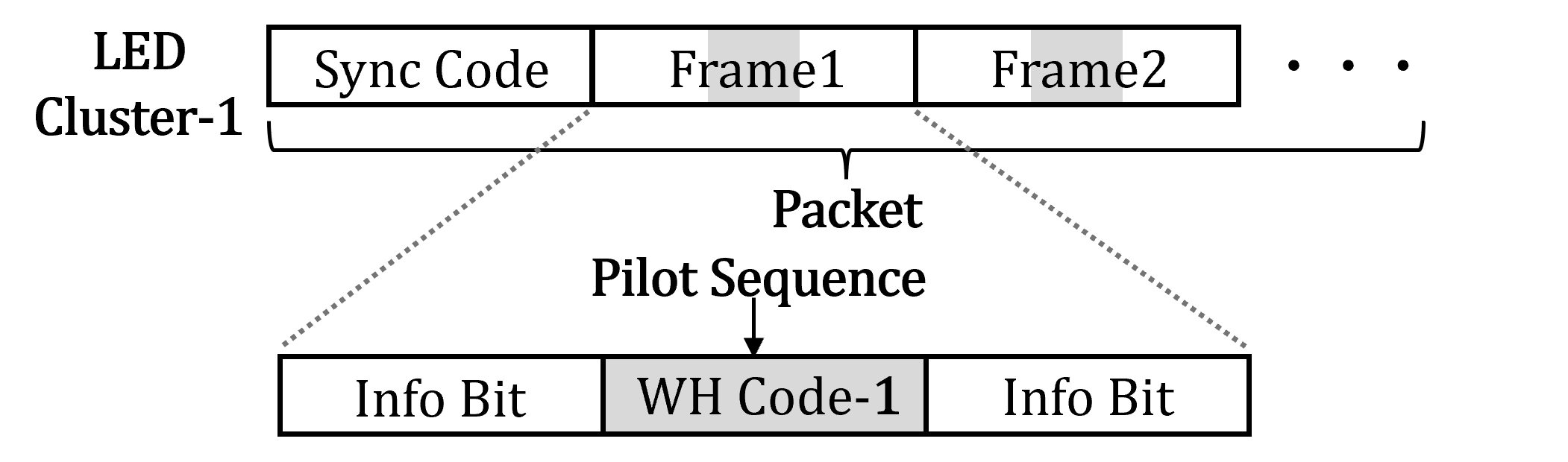}
    \caption{Packet Structure of Transmission Data}
    \vspace{-7mm}
    \label{fig:Transmitter-packet}
\end{figure}

The system model proposed in this study is shown in Fig.~\ref{fig:systemmodel}.
For the transmitter, we used an LED bar consisting of 96 red LEDs aligned vertically, operating at a blinking frequency of 10,000 Hz (Fig.~\ref{fig:rec-and-tra}). These LEDs are grouped into clusters, with each cluster transmitting the same waveform. Grouping LEDs into clusters allows for longer communication distances. In addition, different waveforms are assigned to each LED cluster.
The transmission data is first encoded using Walsh-Hadamard (WH) codes~\cite{Katayama2849}. Next, pilot sequences (known signals unique to each LED cluster) based on WH codes are inserted between the encoded data (information bits). A unit consisting of these information bits and one pilot sequence is referred to as a \emph{frame}. Subsequently, multiple frames are concatenated, and a Barker code is added at the beginning as a synchronization code, forming a \emph{packet}. This packet is then continuously transmitted using On-Off Keying (OOK) modulation from each LED cluster of the LED bar (Fig.~\ref{fig:Transmitter-packet}).

For the receiver, we use a SilkyEvCam equipped with a SONY IMX636 sensor (temporal resolution below 100 µs, resolution 1280$\times$720) (Fig.~\ref{fig:rec-and-tra}). The LED bar is captured using the event camera, and Metavision Studio software is used for data recording. The software applies a high-pass frequency filter to the detected events. As this filtering is performed at the hardware level, the data output from the event camera is significantly reduced.

Initially, the acquired data is processed with an alternation number filter. Then, autocorrelation is performed in the synchronization code portion by shifting time, and the peak position is detected to establish synchronization by identifying the start time of the transmission~\cite{Katayama2867}. Furthermore, the centroid of the LED bar identified by the alternation number filter is calculated, and the displacement of the LED bar between pilot sequences is estimated. Based on this displacement, the data is corrected to account for movement between pilot sequences. Subsequently, cross-correlation with the pilot sequence is performed to determine the position of each LED cluster and separate the transmitted data for each cluster. Finally, the separated data is decoded using Walsh-Hadamard codes.
We will explain the encoder and decoder in the next subsection.

\subsection{Improved Walsh-Hadamard Encoding and Decoding}
\label{method:wh}
In this study, we employ an encoding and decoding method that improves upon techniques proposed in previous research~\cite{Katayama2691, Katayama2849}. Walsh-Hadamard (WH) codes are derived from the rows of a Hadamard matrix, which is an orthogonal matrix. Codes generated from the same matrix exhibit orthogonality, resulting in excellent cross-correlation properties~\cite{Kiyasu1980}. Additionally, each WH code has a unique number of transitions (polarity reversals). However, since event cameras only output events when polarity reversals occur and pixel intensity changes, codes with fewer transitions do not generate events, making communication challenging. Therefore, only codes with a higher number of transitions are used, which reduces the number of codewords and lowers the code rate.

To address this issue, this study introduces a modification by incorporating polarity-inverted WH codes. This approach improves the code rate while maintaining communication performance. Specifically, the system utilizes half of the 16 WH codes generated from a 16×16 Hadamard matrix, along with eight polarity-inverted codes, making a total of 16 codes.

Encoding involves dividing the transmission data into 4-bit units, treating them as 4-digit binary numbers, and converting them into decimal values (0--15). These values are then mapped to the corresponding WH codes for encoding. On the decoding side, the inner product between the received data and all codewords is calculated, and the codeword with the highest similarity (maximum inner product value) is selected as the decoding result.

Event cameras have a limitation where no events are generated if the same brightness value of the LED continues, making waveform reconstruction difficult. However, this method solves the problem by using similarity measures for decoding, even when waveform reconstruction is infeasible. Moreover, the redundancy introduced by this method provides error correction capability. Although polarity-inverted WH codes are used, the cross-correlation properties between the original and inverted WH codes remain below zero. As a result, the decoding process based on the maximum cross-correlation value is not affected. This improvement successfully increases the code rate from 3/16 to 1/4.

\subsection{Positioning and Separating Multiple LED Clusters}
\label{method:clusters}

When using multiple LED clusters, separating them becomes a challenge. The LED bar used in this study has LEDs densely arranged in the vertical direction, making separation through image processing difficult. Furthermore, since this study assumes an in-vehicle environment where the receiver moves, the LEDs shift within the field of view, necessitating tracking of LED clusters. To address these challenges simultaneously, this study proposes a method that utilizes Walsh-Hadamard (WH) codes as pilot sequences.

\begin{figure}[t]
    \centering
    \includegraphics[width = \linewidth]{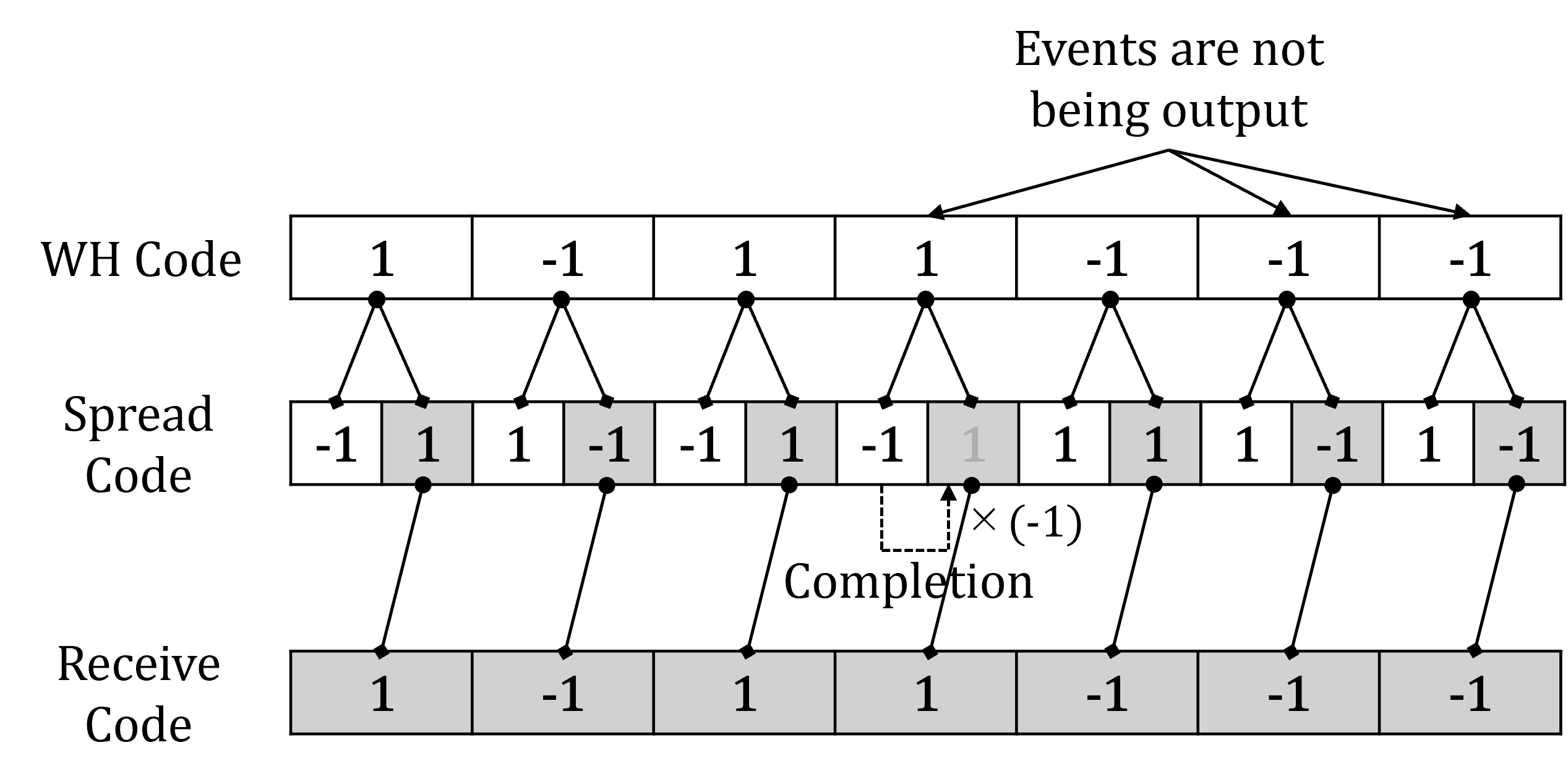}
    \caption{Spread using a bipolar series ([-1,1]) and completion Algorithm }
    \vspace{-4mm}
    \label{fig:Bipolar-Spread}
\end{figure}

On the transmitter side, unique WH codes are assigned to each LED cluster as pilot sequences, and these are spread using a bipolar series ([-1,1]) (see Fig.~\ref{fig:Bipolar-Spread}). The spreading addresses the issue that event cameras, which detect only polarity changes, cannot capture consecutive bits of the same polarity. This characteristic makes it generally difficult to reconstruct transmitted waveforms. 

\begin{figure*}[tbp]
\centering
\includegraphics[width=0.9\linewidth]{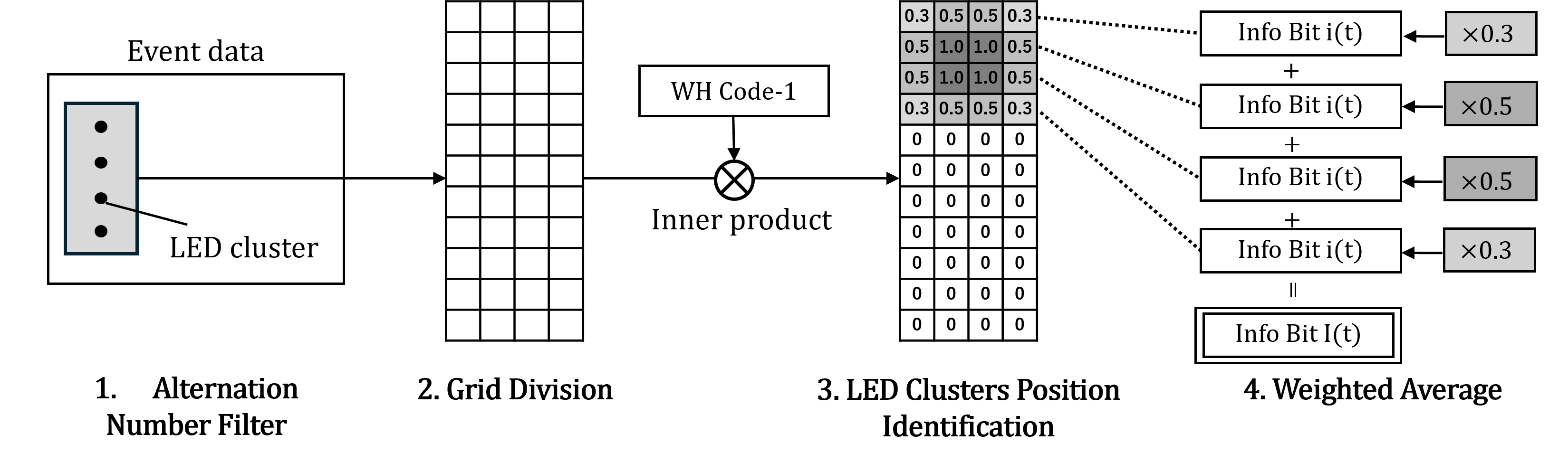}
\caption{LED Cluster Position Identification and Separation Algorithm on the Receiver Side
}
\vspace{-4mm}
\label{fig:Reciever-Algo}
\end{figure*}

In the spread bit sequence, bits multiplied by ``-1'' in the bipolar series alternate with bits multiplied by ``1''.  This sequence is defined as $\mathbf{B} = [b_1, b_2, \dots, b_n]$.
Notably, a polarity reversal always occurs between these two types of bits, allowing waveform reconstruction by using only the bits multiplied by "1" every other bit (\ref{eq:1bitskip}).This is expressed as $\mathbf{B'} = [b'_1, b'_2, \dots, b'_n]$. Furthermore, to address missing events caused by noise, the bits multiplied by "-1," which represent the polarity-reversed original bits, can be inverted and used to compensate for the missing events (\ref{eq:hokan}). This complemented bit sequence is expressed as \(\mathbf{B''} = [b''_1, b''_2, \dots, b''_n]\). Additionally, the WH codes used here are augmented by including the polarity-inverted WH codes, similar to the decoding process. This increases the number of codewords from the original 16 WH codes of sequence length 16 to 32. As a result, these 32 codewords can be assigned to 32 LED clusters.
\begin{equation}
 b'_i = b_{2i} (i = 1,2, \dots, n/2)
 \label{eq:1bitskip}
\end{equation}
\begin{equation}
b''_i =
\begin{cases}
b'_i & \text{if } b'_i \neq 0, \\
-b_{2i-1} & \text{if } b'_i = 0 \text{ and } i > 1,
\end{cases}
\label{eq:hokan}
\end{equation}
The receiver algorithm is shown in Fig.~\ref{fig:Reciever-Algo}. First, an alternation number filter is used to identify the approximate position of the entire LED bar and remove unwanted background events. The alternation number filter operates by filtering based on the number of events per unit time. Since the LEDs are flashing at high speeds, their alternation number is very high, whereas the alternation number of background events is low, allowing for separation~\cite{Katayama2716}.
Subsequently, the vibration-corrected data (details in \cref{method:vibrationcorrection}) is divided into grids, grouping multiple pixels together. Grid processing ensures that data within the same grid can be treated collectively in subsequent steps. Additionally, combining information from multiple pixels improves resistance to electronic noise and missing events.

Next, for the pilot sequence portion $\mathbf{i}_{x,y}(t)$, the inner product with the assigned WH code $N^{(k)}$ is calculated for each grid \eqref{eq:inner}. Grids with high inner product values are considered to have a high probability of containing the corresponding LED cluster, and the inner product values are treated as the existence probability $w_{x,y}$ of the LED cluster. Finally, for grids where the probability exceeds the threshold \(\theta\) \eqref{eq:threshold}, the information bit values \(i_{x,y}(t)\) in each grid are weighted by their respective probabilities and summed to compute the final information bit \(\mathbf{I}^{(k)}(t)\), as shown in \eqref{eq:ave}.
\begin{equation}
    w_{x,y}^{(k)} = \mathbf{i}_{x,y} \cdot \mathbf{N}_k
    \label{eq:inner}
\end{equation}
\begin{equation}
w_{x,y}^{(k)} =
\begin{cases} 
0 & \text{if } w_{x,y}^{(k)} \leq \theta, \\
w_{x,y}^{(k)} & \text{if } w_{x,y}^{(k)} > \theta.
\end{cases}
\label{eq:threshold}
\end{equation}
\begin{equation}
    \mathbf{I}^{(k)}(t) = \sum_{x,y} w_{x,y}^{(k)} \mathbf{i}_{x,y}(t)
\label{eq:ave}
\end{equation}
Unlike the decoding process described in \cref{method:wh}, this method adopts spreading with a bipolar series. This allows waveform reconstruction and maximizes the cross-correlation properties of WH codes. As a result, the inner product values for grids containing other LED clusters are minimized, suppressing interference between LED clusters. This approach enables spatial separation of densely packed LED clusters and tracking even in dynamic environments.

The pilot sequence length required for position identification is 32, corresponding to a transmission time of 3.2 ms. This enables high-resolution temporal tracking. Moreover, grids with high correlation values are more likely to receive correct waveforms for the information bit portion, contributing to the selection of reliable grids. Since this method does not depend on the arrangement of LED clusters, it is applicable not only to vertically aligned LEDs but also to horizontally aligned and other configurations.
\begin{figure}[ht]
    \centering
    \includegraphics[width = 0.8\linewidth]{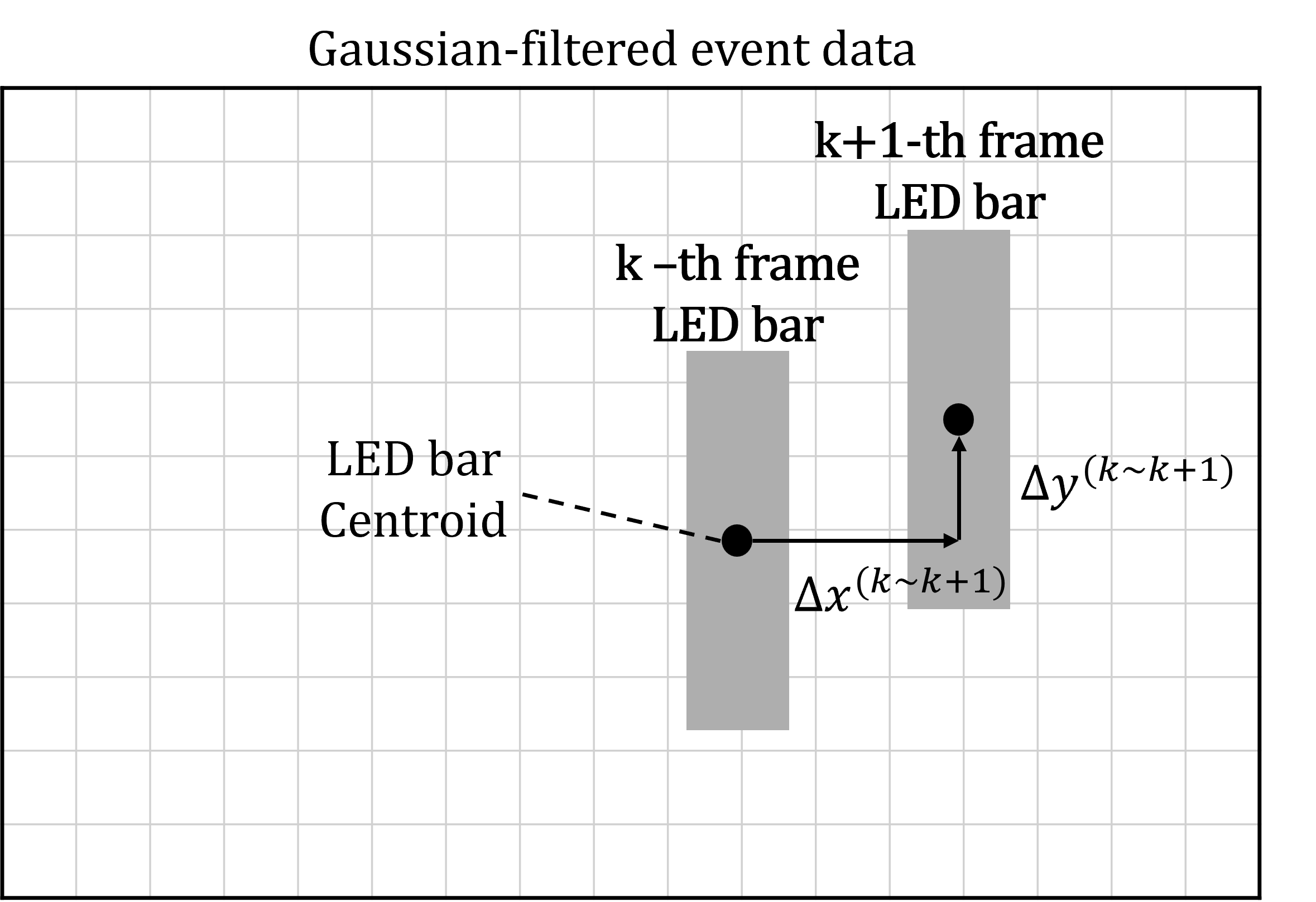}
    \caption{Vibration Detection Algorithm }
    \vspace{-4mm}
    \label{fig:moving}
\end{figure}
\subsection{Vibration Correction Algorithm}
\label{method:vibrationcorrection}
In vehicular environments, the receiver is affected by vibrations caused by the motion of the vehicle. While the algorithm proposed in Section B can track LED clusters, it only identifies their position once per frame (approximately 13 ms). As a result, large and rapid vibrations can cause the tracking to fail to keep up~\cite{Katayama1998}. According to a study using a 35 mm lens and a high-speed camera with 512×512 pixels mounted on a vehicle traveling at 30 km/h (11.1 m/s), vertical movement exceeding 1.5 pixels per millisecond can occur. Additionally, analyses of event camera data revealed that vibrations of approximately 8 pixels per 10 ms were sometimes observed. While it is possible to insert more pilot sequences to address this issue, doing so reduces the proportion of information bits, thereby lowering the communication rate.

To address this issue, this study proposes a method to interpolate the movement of LED clusters between pilot sequences (Fig.~\ref{fig:moving}). First, events corresponding to the pilot sequence portion are aggregated, and an image is generated by recording the number of events for each pixel. A Gaussian filter is then applied to the image to remove noise and smooth out event density variations caused by the LED bar. Subsequently, pixels with event counts exceeding a predefined threshold are selected, and the centroid of these pixels is calculated to determine the center coordinates of the LED bar. This process is performed for the pilot sequences contained in three consecutive frames (the \((i-1)\)-th to \((i+1)\)-th frames). The calculated center coordinates are defined as \((X_p^{(k)}, Y_p^{(k)})\), where \(k = i-1, i, i+1\). Based on the displacement \(\Delta x\) and \(\Delta y\) of the LED bar's center coordinates calculated between two images, corrections are applied to offset the movement from the time the pilot sequence was transmitted. These are expressed as Equations (\ref{eq:tmid})–(\ref{eq:y_correction}). Here, \( t_{i,\text{start}} \) and \( t_{i,\text{end}} \) represent the start and end times of each frame, and \(\alpha(t)\) is the correction factor. Additionally, the coordinate information of the events is represented by \(x\) and \(y\), while the corrected coordinates are represented by \(x'\) and \(y'\).

\begin{equation}
\begin{cases}
 \Delta x^{(k \sim k+1)} = X_p^{(k+1)} - X_p^{(k)} \\
 \Delta y^{(k \sim k+1)} = Y_p^{(k+1)} - Y_p^{(k)}
\end{cases}
\label{eq:movemetn}
\end{equation}
\begin{equation}
t_{mid}= \frac{t_{i, \text{start}} + t_{i, \text{end}}}{2}
\label{eq:tmid}
\end{equation}
\begin{equation} 
\alpha(t) = \frac{t_{mid} - t}{t_{i,end} - t_{i,start}}
\label{eq:alpha} 
\end{equation}
\begin{equation} 
x_j' = x_j + \alpha(t) \cdot 
\begin{cases} \Delta x^{(i-1  \sim  i)}, & t_{i, \text{start}} \leq t < t_{\text{mid}}\\ 
\Delta x^{(i  \sim  i+1)}, & t_{\text{mid}} \leq t \leq t_{i, \text{end}} \end{cases}
\label{eq:x_correction}
\end{equation}
\begin{equation} 
y_j' = y_j + \alpha(t) \cdot 
\begin{cases} \Delta y^{(i-1  \sim  i)}, & t_{i, \text{start}} \leq t < t_{\text{mid}}\\ 
\Delta y^{(i  \sim  i+1)}, & t_{\text{mid}} \leq t \leq t_{i, \text{end}} \end{cases}
\label{eq:y_correction}
\end{equation}
The use of the pilot sequence is crucial because, unlike the information bits, the pilot sequence transmits the same waveform across all frames. This consistency ensures that the shape of the LED bar matches between the two images, leading to improved accuracy.Moreover, this correction method requires very low computational cost, making it suitable for environments where real-time performance is critical.
\section{Field Experiments}
\label{sec:experiment}
\subsection{Experiment in a Static Environment}
The previously described system model was evaluated in a stationary environment with both the transmitter and receiver fixed in place. Randomly generated binary data (0s and 1s) were used as the transmission data. LED clusters were configured with 12, 16, and 32 clusters by grouping 8, 6, and 3 adjacent LEDs, respectively. The communication rates for these configurations were 57 kbps, 28 kbps, and 21 kbps, respectively. The experiment was conducted under clear weather conditions using a 25 mm lens. Both the event camera and the LED bar were securely mounted.

\begin{figure}[t]
    \centering
    \includegraphics[width = \linewidth]{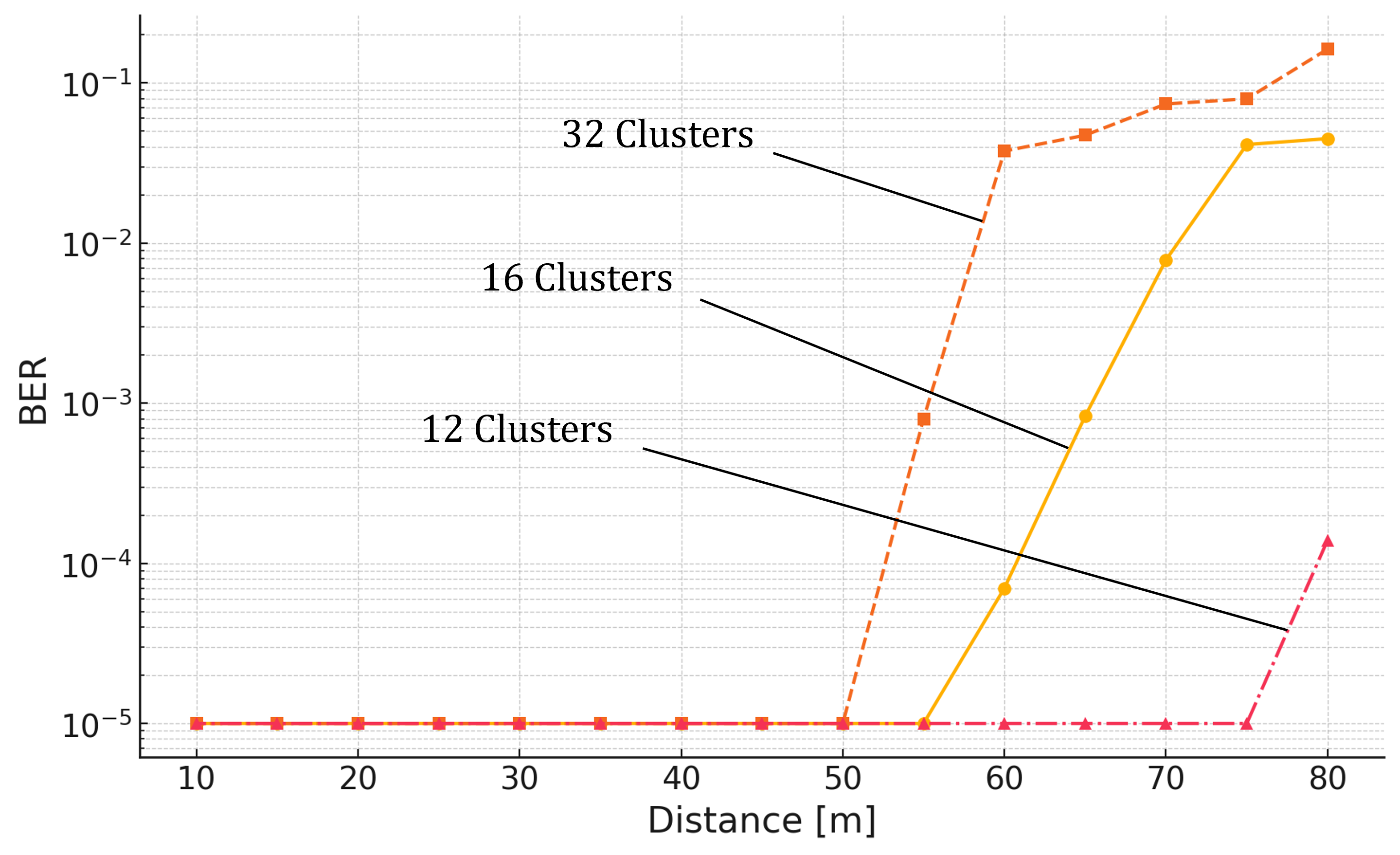}
    \caption{BER-Distance Characteristics in a Stationary Environment}
    \vspace{-4mm}
    \label{fig:stop-exp}
\end{figure}

The results are shown in Fig.~\ref{fig:stop-exp}. Under error-free conditions, the BER was plotted as \(10^{-5}\). From these results, it was confirmed that for the configuration with 16 LED clusters, error-free communication was achieved at distances up to 55 m between the transmitter and receiver.
\subsection{Experiment in a Vehicular Environment}
Next, experiments were conducted with the receiver mounted on a moving vehicle. The event camera was securely attached to the side of the vehicle using a suction-cup tripod. The experiments were conducted at speeds of 20 km/h (5.6 m/s), 30 km/h (8.3 m/s), and 40 km/h (11.1 m/s) to simulate conditions for small mobility vehicles. Based on the results of the stationary environment evaluation, the configuration with 16 LED clusters was selected to balance communication distance and rate.

To measure the distance between the transmitter and receiver, pylons were placed at 5-meter intervals, and an image sensor camera synchronized with the event camera captured footage from inside the vehicle. Using this footage, the exact distances at each time point were calculated.

\begin{figure}[t]
    \centering
    \includegraphics[width = \linewidth]{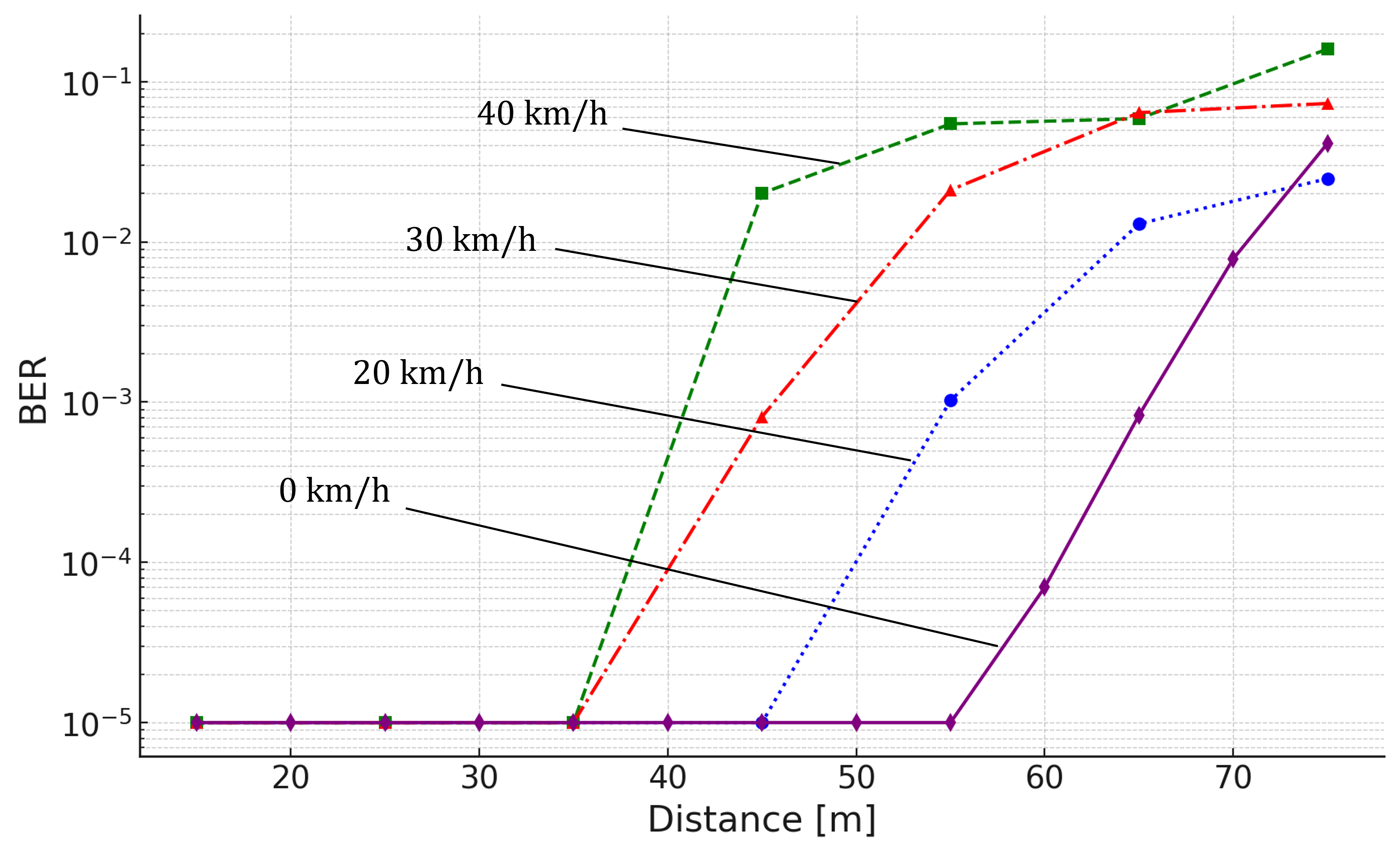}
    \caption{BER-Distance Characteristics in a Vehicular Environment}
    \vspace{-7mm}
    \label{fig:move-exp}
\end{figure}

For analysis, the BER of the data received while the vehicle traveled 10 meters was averaged. Furthermore, the best result from several experiments is shown in Fig.\ref{fig:move-exp}. As a result of the experiment, error-free communication was achieved within a distance of less than 50 meters at 20 km/h (5.6 m/s), less than 40 meters at 30 km/h (8.3 m/s), and less than 40 meters at 40 km/h (11.1 m/s).
These results confirm that LED bar tracking and separation of individual LED clusters were successfully achieved even in a moving environment.
\section{Conclusion}
\label{sec:conclution}
This study proposes a vehicular visible light communication (VLC) system utilizing multiple LEDs and an event camera. The system enables the receiver to separate LED clusters and track them within the field of view in vehicular environments. Notably, it achieves LED position tracking with a high temporal resolution of 3.2 ms. Furthermore, the system is not limited by the geometric arrangement of the LEDs and can be applied to configurations other than the proposed LED bar. Using multiple information sources, it achieves higher communication rates compared to systems using a single information source.

Field trials were conducted to evaluate the system in a vehicular environment. Error-free communication was confirmed under conditions simulating small mobility vehicles, specifically at a speed of 30 km/h (8.3 m/s) and distances within 40 m. These results demonstrate that, compared to VLC systems using image sensors or photodiodes, the proposed system functions as "eyes" for sensing the vehicle's surroundings while achieving a high communication rate. Thus, it represents a promising option as a vehicular sensor.

However, at present, the system lacks sufficient speed adaptability for standard vehicles traveling at 60 km/h. To overcome this limitation, the development of a system capable of higher temporal resolution tracking is necessary. As part of this effort, the design of more efficient pilot sequences will be explored. Additionally, enabling longer-distance communication requires reducing the bit error rate (BER). Introducing powerful error-correcting codes, such as Polar codes, is expected to achieve practical BER levels even for long-distance communication.

\vspace{-1mm}
\section*{Acknoledgement}
We would like to thank Prof. H. Okada (Nagoya Univ.) for their useful discussions.


\begin{thebibliography}{00}
\bibitem{IEEE2407}
{D.Chandra E and K.K.Pedapenki, "Communication and Control for Secure Autonomous Connected Vehicles in V2X Architecture - A Review," in 2023 Fourth International Conference on Smart Technologies in Computing, Electrical and Electronics, December 2023, DOI: 10.1109/ICSTCEE60504.2023.10585213}

\bibitem{IEEE9307}
{E.Cinque ,F.Valentini ,A.Persia ,S.Chiocchio ,F.Santucci and M.Pratesi, "V2X Communication Technologies and Service Requirements for Connected and Autonomous Driving," in  2020 AEIT International Conference of Electrical and Electronic Technologies for Automotive, DOI:10.23919/AEITAUTOMOTIVE50086.2020.9307388}

\bibitem{IEEE2019}
{S.Cho, S.Park, K.M.Kang, S.Ahn, "Analysis of Spectrum Requirements for Autonomous Driving Using SINR Probability Distributions," in IEEE Communications Letters, vol. 24, pp. 202 - 206, Januray 2020, DOI: 10.1109/LCOMM.2019.2952109}

\bibitem{IEEE2300}
{G.Singh, A.Srivastava, V.A.Bohara,M.N.A.Rahim, Z.Liu and D.Pesc, "Towards 6G-V2X: Aggregated RF-VLC for Ultra-Reliable and Low-Latency Autonomous Driving," in IEEE Communications Standards Magazine, Vol 8, Issue 4, December 2024, DOI:10.1109/MCOMSTD.0001.2300005}


\bibitem{IEEE2022}
{A.M.Căilean, C.Beguni, S.A.Avătămăniţei, and M.Dimian, "Experimental Demonstration of a 185 meters Vehicular Visible Light Communications Link," in 2021 IEEE Photonics Conference, October 2021, DOI: 10.1109/IPC48725.2021.9592878}

\bibitem{IEEE2024}
{M.Plattner and G.Ostermayer, "Camera-based Vehicle-to-Vehicle Visible Light Communication - A Software-Only Solution for Vehicle Manufacturers," in 2023 32nd International Conference on Computer Communications and Networks, July 2023, DOI: 10.1109/ICCCN58024.2023.10230125}

\bibitem{Katayama2023}
{T.Yamazato, A.Ohmura, H.Okada, T.Fujii, T.Yendo, S.Arai and K.Kamakura, "Range estimation scheme for integrated I2V-VLC using a high-speed image sensor," in IEEE International Conference on Communications Workshops, pp.326 - 330, May 2016, DOI: 10.1109/ICCW.2016.7503808}



\bibitem{Lichtsteiner2006isscc}
{P.Lichtsteiner, C.Posch, and T.Delbruck, ``A 128x128 120dB 30mW asynchronous vision sensor that responds to relative intensity change,'' In IEEE Int. Solid-State Circuits Conf. (ISSCC), pages 2060–2069, 2006}

\bibitem{Gallego2022pami}
{G.Gallego, T.Delbruck, G.Orchard, C.Bartolozzi, B.Taba, A.Censi, S.Leutenegger, A.Davison, J.Conradt, K.Daniilidis, and D.Scaramuzza, ``Event-based vision: A survey,'' in IEEE Trans. Pattern Anal. Mach. Intell., 44(1):154–180, 2022}

\bibitem{Gehrig2024nature}{
D.Gehrig, and D.Scaramuzza, ``Low-latency automotive vision with event cameras,'' Nature 629, 1034–1040 (2024). DOI: 10.1038/s41586-024-07409-w}

\bibitem{IEEE2018}
{W.H.Shen, P.W.Chen and H.M.Tsai, "Vehicular Visible Light Communication with Dynamic Vision Sensor: A Preliminary Study," in 2018 IEEE Vehicular Networking Conference, December 2018, DOI: 10.1109/VNC.2018.8628425}

\bibitem{ACM2019}
{W.H.Shen, "Vehicular Visible Light Communication with Dynamic Vision Sensor" in RisingStarsForum'19: The ACM MobiSys 2019 on Rising Stars Forum, pp. 25 - 30, June 2019, DOI: 10.1145/3325425.3329941 }

\bibitem{Arxiv2024}
{Y.Wang, Y.Shen, K.Xu, M. Hassan, G.Zhao, C. Xu and W. Hu, "Towards High-Speed Passive Visible Light Communication with Event Cameras and Digital Micro-Mirrors" in 22nd ACM Conference on Embedded Networked Sensor Systems, pp. 704 - 717, November 2024, DOI:10.1145/3666025.369936}

\bibitem{Katayama2849}
{D.Ehara, "Event-based Vision Sensor Visible Light Communication using Walsh-Hadmard Codes to Separate Background Noise," Master's thesis, Graduate school of Engineering, Nagoya University, March 2024}

\bibitem{Katayama2867}
{R.Soga, S.Lu and T.Yamazato, "Synchronization method using Barker code in Event-Based-Camera Visible Light Communication using Walsh-Hadamard code," in Enginnering Sciences Society Conference of IEICE, A-9-7, September 2024}

\bibitem{Katayama2691}
{D.Ehara, Z.Tang, M.Kinoshita, T.Yamazato, H.Okada, K.Kamakura, S.Arai, T.Yendo and T.Fujii, "Influence of Walsh-Hadamard Code Sequency in Visible Light Communication Using an Event Camera," in International Conference on Emerging Technologies for Communications, S9-5,, December 2022, DOI: 10.34385/proc.72.S9-5}

\bibitem{Kiyasu1980}
{Z. Kiyasu, "Hadamard Matrices and Their Applications," IEICE Transactions, 1980.}

\bibitem{Katayama2716}
{K.Furukawa, D.Ehara, T.Yamazato, H.Okada, M.Kinoshita, T.Yendo, S.Arai, K.Kamakura and T.Fujii, "Analysis of Event Occurrence under the Moving Situation of the Event-Based Camera," in IEICE General Conference, A-9-3, pp.82, March 2023}

\bibitem{Katayama1998}
{M.Kinoshita, T.Yamazato, H,Okada, T.Fujii, S.Arai, T.Yendo, K.Kamakura, "Channel Fluctuation Measurement for I2V-VLC, V2I-VLC, and V2V-VLC Using Image Sensor" in IEICE Technical Report, ITS2014-18, pp. 77 - 81, September 2014}

\end{thebibliography}
\end{document}